\documentclass{article}
\usepackage{spconf,amsmath,graphicx}
\usepackage{multirow}
\usepackage{amssymb}
\usepackage{array}
\usepackage{color}

\title{Multiscale Domain Adaptive YOLO for Cross-Domain Object Detection}
\name{Mazin Hnewa and Hayder Radha}
\address{
Michigan State University, 
East Lansing, MI 48824,
United States\\
{\tt\small (mazin,radha)@msu.edu}
}
\begin{document}
%
\maketitle
\begin{abstract}
The area of domain adaptation has been instrumental in addressing the domain shift problem encountered by many applications. This problem arises due to the difference between the distributions of source data used for training in comparison with target data used during realistic testing scenarios. In this paper, we introduce a novel MultiScale Domain Adaptive YOLO (MS-DAYOLO) framework that employs multiple domain adaptation paths and corresponding domain classifiers at different scales of the recently introduced YOLOv4 object detector to generate domain-invariant features. We train and test our proposed method using popular datasets. Our experiments show significant improvements in object detection performance when training YOLOv4 using the proposed MS-DAYOLO and when tested on target data representing challenging weather conditions for autonomous driving applications.
\end{abstract}

\section{Introduction}
\label{sec:intro}
\begin{figure}[t]
\begin{center}
   \includegraphics[width=1\linewidth]{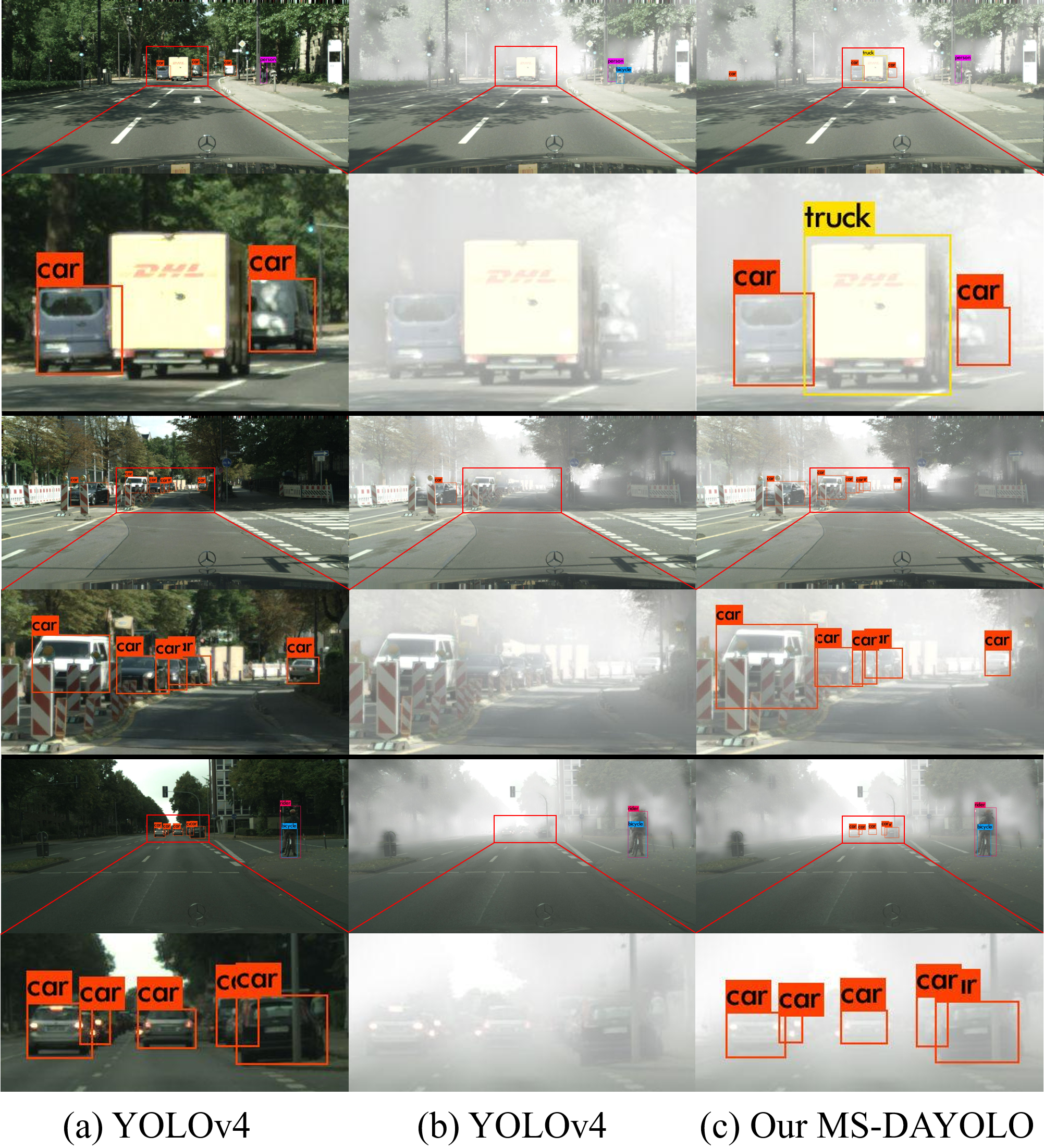} 
\end{center}
  \caption{Visual detection examples using the original YOLOv4 method on: (a) clear images and (b) foggy images. (c) Our proposed MS-DAYOLO applied onto foggy images.} 
\label{fig:cityscapes_examples}
\end{figure}

Convolutional Neural Networks (CNNs) have been achieving exceedingly improved performance for object detection in terms of classifying and localizing a variety of objects in a scene \cite{faster_RCNN,ssd,yolo,RetinaNet}. However, under a domain shift, when the testing data has a different distribution from the training data distribution, the performance of state-of-the-art object detection methods drop noticeably and sometimes significantly. Such domain shift could occur due to capturing the data under different lighting or weather conditions, or due to viewing the same objects from different view points leading to changes in object appearance and background. For example, training data used for autonomous vehicles is normally captured under favorable clear weather conditions whereas testing could take place under more challenging weather (\textit{e.g.} rain, fog). Consequently, methods fail to detect objects as shown in the examples of Figure \ref{fig:cityscapes_examples}(b). In that context, the domain under which training is done is known the \textit{source domain} while the new domain under which testing is conducted is referred to as the \textit{target domain}.

One of the challenges that aggravates the domain shift problem is the lack of annotated target domain data. This led to the emergence of the area of \textit {domain adaptation} \cite{duan2012domain,kulis2011you,tzeng2017adversarial,long2016unsupervised}, which has been widely studied to solve the problem of domain shift without the need to annotate data for new target domains. 
Recently, domain adaptation has been used to improve the performance of object detection due to domain shift \cite{survey}. It attempts to learn a robust object detector using labeled data from the source domain and unlabeled data from the target domain. Most domain adaptation approaches in literature employ adversarial training strategy \cite{GANs}. In particular, a domain classifier is optimized to identify whether a data point from the source or target domain, while the feature extractor of object detector is optimized to confuse the domain classifier. This strategy makes the feature extractor learn domain invariant features. Domain adaptive Faster R-CNN \cite{DA_faster_RCNN} is the first work that employed adversarial training for domain adaptation based object detection. After that, many adversarial-based methods were developed for domain adaptation object detection \cite{zhu2019adapting,wang2019few,saito2019strong,he2019multi}.


Equally important, the particular domain adaptation solution used is influenced greatly by the underlying object detection method architecture. In that context, within the area of object detection, domain adaptation has been studied rather extensively for Faster R-CNN object detection and its variants \cite{DA_faster_RCNN,zhu2019adapting,wang2019few,saito2019strong,he2019multi}. 
Despite its popularity, Faster R-CNN suffers from long inference time to detect objects. As a result, it is arguably not the optimal choice for time-critical, real-time applications such as autonomous driving. On the other hand, one-stage object detectors, and in particular YOLO, can operate quite fast, even much faster than real-time, and this makes them invaluable for autonomous driving and similar time-critical applications. Furthermore, domain adaptation for the family of YOLO architectures have received virtually no attention. Besides the computational advantage of YOLO, the latest version, YOLOv4, has many salient improvements and its object detection performance has improved rather significantly relative to prior YOLO architectures and more important in comparison to Faster R-CNN. All of these factors motivated our focus on the development of a new domain adaptation framework for YOLOv4.


In this paper, we propose a novel
MultiScale Domain Adaptive YOLO (MS-DAYOLO) that supports domain adaptation at different layers of the feature extraction stage within the YOLOv4 backbone network. In particular, MS-DAYOLO includes a Domain Adaptive Network (DAN) with multiscale feature inputs and multiple domain classifiers.  We conducted extensive experiments using popular datasets. These experiments show that our proposed MS-DAYOLO framework provides significant improvements to the performance of YOLOv4 when tested on target domain as shown in the examples of Figure \ref{fig:cityscapes_examples}(c). To the best of our knowledge, this is the first proposed work that improves the performance of YOLO for cross domain object detection. 

\section{Proposed MS-DAYOLO}

\begin{figure*}
\begin{center}
\includegraphics[width=1\linewidth]{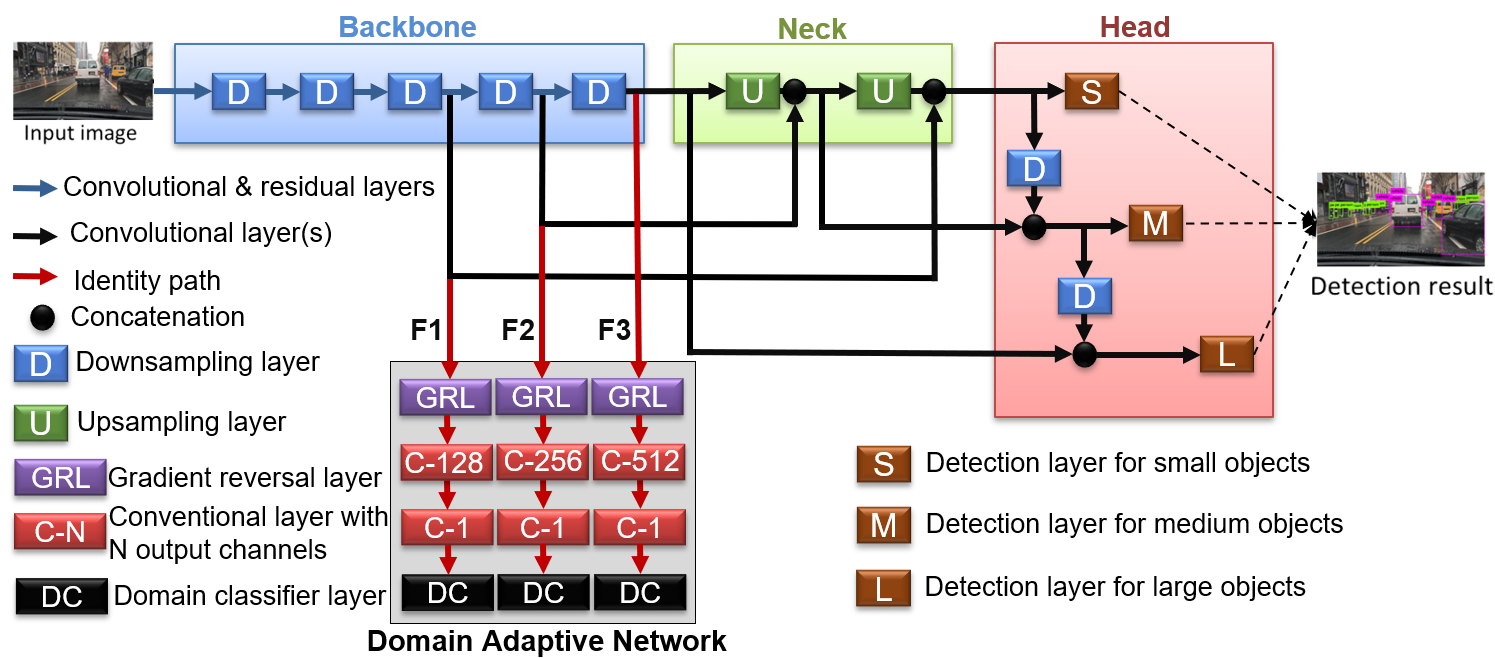} 

\end{center}
 \caption{Architecture of the proposed MS- DAYOLO. Domain adaptation network (DAN) is attached to the YOLO object detector only during training in order to learn domain invariant features.}
\label{fig:arch}

\end{figure*}

YOLOv4 \cite{yolov4} has been released recently as the latest version of the family of the YOLO object detectors. Relative to its predecessor, YOLOv4 has incorporated many new revisions and novel techniques to improve the overall detection accuracy. YOLOv4 has three main parts: backbone, neck, and head as shown in Figure \ref{fig:arch}. The backbone is responsible for extracting multiple layers of features at different scales. The neck collects these features from three different scales of the backbone using upsampling layers and feed them to the head. Finally, the head predicts bounding boxes surrounding objects as well as class probabilities associated with each bounding box.

The backbone (\textit{i.e.}{feature extractor}) represents a major module of the YOLOv4 architecture, and we believe that it makes a significant impact on the overall performance of the detector. In addition to many convolutional layers, it has 23 residual blocks \cite{ResNet}, and five downsampling layers to extract critical layers of features that are used by the subsequent detection stages. Here, we consecrate on the features (F1, F2, and F3 in Figures \ref{fig:arch}) because they are fed to the next stage (neck module). In particular, our goal is to apply domain adaptation to these three features to make them robust against domain shifts at different scales, and hence, have them converge toward domain invariance during domain-adaptation based training. 

\subsection{Domain Adaptive Network for YOLO}

The proposed Domain Adaptive Network (DAN) is attached to the YOLOv4 object detector only during training in order to learn domain invariant features. Indeed, YOLOv4 and DAN are trained in an end-to-end fashion. For inference, and during testing, domain-adaptive trained weights are used in the original YOLOv4 architecture (without the DAN network). Therefore, our proposed framework will not increase the underlying detector complexity during inference, which is an essential factor for many real-time applications such as autonomous driving.

DAN uses the three distinct scale features of the backbone that are fed to the neck as inputs. It has several convolutional layers to predict the domain class (either source or target). Then, domain classification loss ($\mathcal{L}_{dc}$) is computed via binary cross entropy as follows:
 
\begin{equation}
\label{dc_loss}
 \mathcal{L}_{dc} = -\sum_{i,x,y} [ t_i \ln p_i^{(x,y)} + (1-t_i) \ln (1-p_i^{(x,y)})]
 \end{equation}
Here, $t_i$ is the ground truth domain label for the i-th training image, with $t_i=1$ for source domain and $t_i=0$ for target domain. $p_i^{(x,y)}$ is predicted domain class probabilities for i-th training image at location $(x,y)$ of the feature map.

DAN is optimized to differentiate between the source and target domains by minimizing this loss. On the other hand, the backbone is optimized to maximize the loss to learn domain invariant features. Thus, features of the backbone should be indistinguishable for the two domains. Consequently, this should improve the performance of object detection for the target domain. 

To solve the joint minimization and maximization problem, we employ adversarial leaning strategy\cite{GANs}. In particular, we achieve this contradictory objectives by using a Gradient Reversal Layer (GRL) \cite{grl,grl_J} between the backbone and the DAN network. GRL is a bidirectional operator that is used to realize two different optimization objectives. In the feed-forward direction, the GRL acts as an identity operator. This leads to the standard objective of minimizing the classification error when performing local backpropagation within DAN. On the other hand, for backpropagation toward the backbone network, the GRL becomes a negative scalar ($\lambda$). Hence, in this case, it leads to maximizing the binary-classification error; and this maximization promotes the generation of domain-invariant features by the backbone.

To compute the detection loss ($\mathcal{L}_{det}$) \cite{yolo}, only source images are used because they are annotated with ground-truth objects. Consequently, all three parts of YOLOv4 (\textit{i.e.} backbone, neck and head) are optimized via minimizing $\mathcal{L}_{det}$. On the other hand, both source labeled images and target unlabeled images are used to compute the domain classification loss  ($\mathcal{L}_{dc}$) which is used to to optimize DAN via minimizing it, and the backbone via maximizing it. As a result, both $\mathcal{L}_{det}$ and $\mathcal{L}_{dc}$ are used to optimize the backbone. In other words, the backbone is optimized by minimizing the following total lose:

\begin{equation}
\label{loss_backnone}
 \mathcal{L}_t = \mathcal{L}_{det} + \lambda \mathcal{L}_{dc}
 \end{equation}
 where $\lambda$ is a negative scalar of GRL that balances a trade-off between the detection loss and domain classification loss. In fact, $\lambda$ controls the impact of DAN on the backbone.
 
 \subsection{DAN Architecture}

 Instead of applying domain adaptation for only the final scale of the feature extractor as done in the Domain Adaptive Faster R-CNN architecture \cite{DA_faster_RCNN}, we develop domain adaptation for three scales separately to solve gradient vanishing problem. In other word, applying domain adaptation only to the final scale (F3) does not make significant impact to the previous scales (F1 and F2) due to gradient vanishing problem as there are many layers between them. As a result,  we employ a multiscale strategy that connects the three features F1, F2, and F3 of the backbone to the DAN through three corresponding GRLs as shown in Figure \ref{fig:arch}. For each scale, there are two convolutional layers after GRL, the first one reduces the feature channels by half, and the second one predict the domain class probabilities. Finally, a domain classifier layer is used to compute the domain classification loss. 


\section{Experiments}

In this section, we evaluate our proposed MS-DAYOLO. We modified the official source code of YOLOv4 that is based on the darknet platform\footnote{https://github.com/AlexeyAB/darknet}; and we developed new code to implement our proposed method\footnote{https://github.com/Mazin-Hnewa/MS-DAYOLO}.

\subsection{Setup}

For training, we used the default settings and hyper-parameters that were used in the original YOLOv4 \cite{yolov4}. The network is initialized using the pre-trained weights file. The training data includes two sets: source data that has images and their annotations (bounding boxes and object classes), and target data without annotation. Each batch has 64 images, 32 from the source domain and 32 from target domain. Based on prior works \cite{DA_faster_RCNN, zhu2019adapting,he2019multi} and our experience, we set $\lambda =0.1$ for all experiments. 

For evaluation, we report Average Precision (AP) for each class as well as  mean average precision (mAP) with threshold of 0.5 \cite{2011pascal} using testing data that has labeled images of target domain. We have followed other prior domain adaptive object-detection works that use the same threshold value of 0.5. In an extended future version of this work we plan to study the impact of other threshold values. We compare our proposed method with the original YOLOv4, both applied to the same target domain validation set. It is worth noting that most prior domain-adaptive object detection methods used Faster R-CNN as the baseline object detector \cite{DA_faster_RCNN,zhu2019adapting,wang2019few,saito2019strong,he2019multi}. Consequently, we do not compare our proposed method, which is based on the YOLOv4 detector, with those Faster R-CNN based methods. Such comparison would imply comparing different frameworks with completely different underlying-architectures for object detection.

Domain shifts due to changes in weather conditions is one of the most prominent reasons for discrepancy between the source and target domains. Reliable object detection systems in different weather conditions are essential for many critical applications such an autonomous driving. As a result, we focus on presenting our evaluation results of our proposed MS-DAYOLO by studying domain shifts under changing weather condition for autonomous driving. To achieve this, we use different driving datasets: Cityscapes \cite{cityscapes}, Foggy Cityscapes \cite{Foggy_cityscapes}, BDD100K \cite{bdd100k}, and INIT \cite{init}.

\subsection{Results and Discussion}

\textbf{Clear $\rightarrow$ Foggy}: we  discuss the ability of our proposed method to adapt from clear to foggy weather as has been done by many recent works in this area \cite {DA_faster_RCNN,zhu2019adapting,wang2019few,saito2019strong,he2019multi}. Original YOLOv4 is trained using the Cityscapes training set. While MS-DAYOLO is trained using the Cityscapes training set as source domain and the Foggy Cityscapes training set without labels as target domain. The Foggy Cityscapes validation set is used for testing and evaluation. Because Foggy Cityscapes training set is annotated, we are able to train the original YOLOv4 with this set to show the ideal performance (oracle). The Cityscapes dataset has eight classes. However, because the number of ground-truth objects for some classes (truck, bus, and train) is small (\textit{i.e.} less than 500 in training set, and 100 in testing set), the performance measure will be inaccurate for these classes. As a result, we exclude them and compute mAP based on the remaining classes.

To show the important of applying domain adaptation to three distinct scales of the backbone network, we conducted an ablation study. First, we applied domain adaptation, separately, to each of the three scales of features that are fed into the neck of the YOLOv4 architecture. Then we apply domain adaptation to different combinations of two scales at a time. Finally, we compared the results with the performance of applying these combinations of the study with the performance of applying our MS-DAYOLO to all three scales.

Table \ref{mAP_cityscapes} summarizes the performance results. It is clear that based on these results, we can conclude that applying domain adaptation to all three feature scales improves the detection performance on target domain, and achieves the best result. Moreover, our proposed MS-DAYOLO outperforms the original YOLOv4 approach by significant margin, and it almost reaches the performance of the ideal (oracle) scenario, especially for some object classes in terms of average precision and overall mAP. Figure \ref{fig:cityscapes_examples} shows examples of detection results of the proposed method as compared to the original YOLOv4.     

\begin{table}[!t]
\centering
\caption{Quantitative results on adaptation from clear to foggy weather of the Cityscapes dataset. \checkmark means that domain adaptation is applied to the feature scale(s) using our MS-DAYOLO. The classes are P:Person, R:Rider, C:Car, M: Motorcycle, and B: Bicycle. Results in red are obtained using the baseline YOLO for comparison with our method.}
\label{mAP_cityscapes}
\begin{tabular} {|m{2.6em}|m{0.1em}m{0.1em}m{0.5em}|m{1.4em}m{1.4em}m{1.4em}m{1.4em}m{1.7em}|m{1.7em}|}
\hline
 Method & F1 & F2 & F3 & P & R
 & C & M & B & mAP  \\
\hline
 YOLO & &  &  & \textcolor{red}{31.57}& \textcolor{red}{38.27} & \textcolor{red}{46.93} & \textcolor{red}{16.75}	& \textcolor{red}{30.32} & \textcolor{red}{32.77}\\
\hline

\multirow{7}{*} {Ours} & & & \checkmark & 36.84&	42.84&	53.69&		24.77&	32.35 & 38.10\\ 
\cline{2-10}

 & &\checkmark  &  &	37.08&	41.49 &	54.49&		26.22&	32.43 & 38.34\\
 
\cline{2-10}
 & \checkmark &  &  &	36.28 &	44.22 & 53.10 &		25.81&	35.87 & 39.06 \\
 
\cline{2-10}

 & & \checkmark& \checkmark & 36.62&	42.68&	55.70 &  26.09 &	33.52& 38.92  \\
\cline{2-10}

& \checkmark & \checkmark&  &37.50 &	42.48 &	54.53  &27.84 &	34.75 &	39.42 \\
\cline{2-10}
&\checkmark & & \checkmark &  36.41&	\textbf{46.06}&	52.19& 22.48 &34.99 &38.43\\
\cline{2-10}

&\checkmark & \checkmark& \checkmark &\textbf{38.62}	 &45.52&	\textbf{55.85} & \textbf{28.82} &	\textbf{36.46} &	\textbf{41.05}\\
\hline
\hline
 \multicolumn{4}{|l|}{Oracle} & 42.35 & 49.50  &	63.59   &	31.10 &	39.68 &	45.24  \\
\hline

\end{tabular}
\end{table}

\textbf{Sunny $\rightarrow$ Rainy}:  we also discuss the ability of our proposed method  to adapt from sunny to rainy weather using BDD100K \cite{bdd100k} and INIT \cite{init} datasets. We extracted "sunny weather" labeled images for the source data, and  "rainy weather" unlabeled images to represent the target data. As before, the original YOLOv4 is trained using only source data (\textit{i.e.} labeled sunny images). Meanwhile, our proposed MS-DAYOLO is trained using both source and target data (\textit{i.e.} labeled sunny images and unlabeled rainy images). In addition, we extracted labeled images from the rainy-weather data for testing and evaluation. The results are summarized in Table \ref{mAP_rainy}. A clear performance improvement is achieved by our method over the original YOLO in both datasets.

\begin{table}[!t]
\centering
\caption{Quantitative results on adaptation from sunny to rainy weather of the BDD100K and INIT datasets. The classes of BDD100K are V: Vehicle, P:Person, TS:Traffic Sign, and TL:Traffic Light. The classes of INIT are P:Person, C: Car, and SLS: Speed Limit Sign.}
\label{mAP_rainy}

\begin{tabular} 
{|c|m{2.9em}m{2.9em}m{2.8em}m{2.8em}|c|}

\hline
 \multicolumn{6}{|c|}{BDD100K}  \\
\hline
 Method & V & P & TS & TL & mAP  \\
\hline
 YOLO   & 72.54 & 41.54 &56.06 &47.07& 54.30  \\
\hline

Ours  & \textbf{73.74} & \textbf{45.37} & \textbf{58.32} & \textbf{48.00}& \textbf{56.36} \\ 

\end{tabular}
\vspace{-0.45cm}
\end{table}
\begin{table}[!t]
\centering
\label{mAP_init}

\begin{tabular} {|c|m{4.2em}m{4.2em}m{4.2em}|c|}

\hline
\hline
 \multicolumn{5}{|c|}{INIT}  \\
\hline
 Method & P & C & SLS & mAP\\
\hline

 YOLO   &  44.52&74.48&48.39& 55.80 \\
\hline

Ours  & \textbf{48.80} &\textbf{76.03} & \textbf{50.00} & \textbf{58.28} \\ 
\hline

\end{tabular}
\end{table}


\section{Conclusion}

In this paper, we proposed a multiscale domain adaptation framework for the popular real time object detector YOLO.  Specifically, under our MS-DAYOLO, we applied domain adaptation to three different scale features within the YOLO feature extractor that are fed to the next stage. The proposed method improves the overall detection performance under the target domain because it produces robust domain invariant features that reduce the impact of domain shift. Based on various experimental results, our framework can successfully adapt YOLO to target domains without any need for annotation. Furthermore, the proposed MS-DAYOLO outperformed state-of-the-art YOLOv4 under diverse testing scenarios for autonomous driving applications.

\textbf{Acknowledgement:} This work has been supported by the Ford Motor Company under the Ford-MSU Alliance Program. 
%

\bibliographystyle{IEEEbib}
\bibliography{refs}
\end{document}